\documentclass[letterpaper, 10 pt, conference]{ieeeconf}  % Comment this line out if you need a4paper

\IEEEoverridecommandlockouts                              % This command is only needed if 
\usepackage{amsmath} % assumes amsmath package installed
\usepackage{amssymb}  % assumes amsmath package installed
\usepackage{tabularx}
\usepackage[utf8]{inputenc}
\usepackage[T1]{fontenc}
\usepackage{textcomp}
\usepackage{gensymb}
\usepackage{multirow}

\usepackage[dvipsnames]{xcolor}
\usepackage{graphicx}
\graphicspath{ {./images/}}
\usepackage{float}
\usepackage{capt-of}
\usepackage{etoolbox}
\usepackage{adjustbox}
\usepackage{hyperref}
\usepackage{subcaption} 

\usepackage{cite}

\title{\textbf{Multi-object Monocular SLAM for Dynamic Environments}}
\vspace{-0.2cm}
\author{Gokul B. Nair$^{1}$, Swapnil Daga$^{1}$, Rahul Sajnani$^{1}$, Anirudha Ramesh$^{1}$, Junaid Ahmed Ansari$^{2}$, \\ Krishna Murthy Jatavallabhula$^{3}$, and K. Madhava Krishna$^{1}$% <-this % stops a space
    \thanks{$^1$ Robotics Research Center, KCIS, IIIT Hyderabad, India. {\tt\small gokulbnr@gmail.com}}
    \thanks{$^2$ Currently at Embedded Systems and Robotics, TCS Innovation Labs, Kolkata, India.}
    \thanks{$^3$ Mila, Universite de Montreal, Canada.}
}

\def\etal{\emph{et al.}}
\def\cf{\emph{cf.}}

\begin{document}

\makeatletter
\let\@oldmaketitle\@maketitle
\renewcommand{\@maketitle}{\@oldmaketitle

\centering
% \includegraphics[width=\linewidth]{images/Teaser2.png}%,height=6cm]{images/Teaser.png}
% \vspace{-0.1cm}
% \captionof{figure}{Some results showcasing the efficacy of the proposed monocular object localization system. The system is capable of estimating the shape and pose (without scale-factor ambiguity) of objects located on surfaces that do not share the same plane with the moving monocular camera. The images of the scenes contain the projection of the estimated shapes (wireframes) of cars. On the top of each car, we indicate the distance of the car from the camera (in meters). To the right side of each scene, lies the visualization of the estimated wireframe and road points in 3D. For the first and third scenes, we visualize the wireframes with their respective ground truth 3D bounding boxes (shown in red) on the right, highlighting the accurate localization of the objects. In the second scene, we show the accurately estimated cars in 3D, overlayed on a dense ground truth 3D point cloud. Even the objects at over 50 meters distance on steep slopes are accurately localized.}}
% \label{fig:teaser}
}
\makeatother

\maketitle
% \thispagestyle{empty}
% \pagestyle{empty}

%%%%%%%%%%%%%%%%%%%%%%%%%%%%%%%%%%%%%%%%%%%%%%%%%%%%%%%%%%%%%%%%%%%%%%%%%%%%%%%%
\vspace{-0.5cm}

\begin{abstract}
In this paper, we tackle the problem of \emph{multibody} SLAM from a monocular camera. The term \emph{multibody}, implies that we track the motion of the camera, as well as that of other dynamic participants in the scene.
The quintessential challenge in dynamic scenes is unobservability: it is not possible to \emph{unambiguously} triangulate a moving object from a moving monocular camera.
% Hence, multibody monocular SLAM in dynamic environments remains a long-standing challenge in terms of perception and state estimation.
% Although theoretical solutions exist, practice lags behind, predominantly due to the lack of robust perceptual and predictive models of dynamic participants.
Existing approaches solve restricted variants of the problem, but the solutions suffer \emph{relative scale ambiguity} (i.e., a family of infinitely many solutions exist for each pair of motions in the scene).
% Under restrictions of object motion the problem can be solved, however even here one is entailed to solve for the single family solution to the relative scale problem.  
% The relative scale problem exists since the dynamic objects that get reconstructed with the monocular camera have a different scale when compared to that in which the stationary scene is reconstructed. 
We solve this rather intractable problem by leveraging single-view metrology, advances in deep learning, and category-level shape estimation.
We propose a multi pose-graph optimization formulation, to resolve the relative and absolute scale factor ambiguities involved.
% through an object SLAM pipeline.
% Further, we lift the ego vehicle trajectory obtained from Monocular ORB-SLAM also into metric scales making use of ground plane features thereby resolving the relative scale problem. 
% We present a multi pose-graph optimization formulation to estimate the pose of dynamic objects in the environment. 
This optimization helps us reduce the average error in trajectories of multiple bodies over real-world datasets, such as KITTI~\cite{kitti}.
To the best of our knowledge, our method is the first \emph{practical} monocular multi-body SLAM system to perform \emph{dynamic} multi-object and ego localization in a \emph{unified framework} in \emph{metric} scale.

% Multi-body SLAM in dynamic environments remains a long-standing challenge in terms of perception and state estimation. Although theoretical solutions exist, pratice lags behind, predominantly due to the lack of robust perceptual and predictive models of dynamic participants. We present a data-driven approach to learn motion and observation models for dynamic objects. These learnt models enable us to cast the multi-body visual SLAM problem into a well-known multi pose-graph optimization formulation. We additionally demonstrate that, by making use of \emph{commonsense} priors, one can overcome scale factor ambiguity, touted as the Achilles heel of monocular SLAM.
\end{abstract}

%%%%%%%%%%%%%%%%%%%%%%%%%%%%%%%%%%%%%%%%%%%%%%%%%%%%%%%%%%%%%%%%%%%%%%%%%%%%%%%%

\section{Introduction}
\label{sec:introduction}

Monocular SLAM research has significantly matured over the last few decades, resulting in very stable \emph{off-the-shelf} solutions~\cite{orb,lsd,7782863}. However, dynamic scenes still pose unique challenges for even the best such solutions. In this work, we tackle a more general version of the monocular SLAM problem in dynamic environments: \textbf{\emph{multi-body visual SLAM}}. While monocular SLAM methods traditionally track the ego-motion of a camera and discard dynamic objects in the scene, multi-body SLAM deals with the \emph{explicit} pose estimation of multiple dynamic objects (dynamic \emph{bodies}), which finds important applications in the context of autonomous driving.
% Accurate pose estimation and tracking of multiple objects could have several applications. For example, a car driving on a road will find it beneficial to keep tracking other traffic participants\cite{sivaraman2013looking}. 

% Concretely, \textbf{multi-body monocular SLAM} involves the estimation of the trajectories of multiple freely moving objects undergoing rigid transform using observations purely from a freely moving monocular camera.

Despite being an extremely useful problem, multibody visual SLAM has not received comparable attention to its \emph{uni-body} counterpart (i.e., SLAM using stationary \emph{landmarks}). This can primarily be attributed to the \emph{ill-posedness} of monocular multibody Structure-from-Motion~\cite{multibody-in-practice}. While the scale factor ambiguity of monocular SLAM is well-known \cite{monoslam,orb,lsd,7782863}, the lesser-known-yet-well-studied \emph{relative scale} problem persists with multibody monocular SLAM~\cite{kundu2011,multibody_generic,multibody-in-practice,two-view-multibody,two-view-outlier,two-view-robust, Namdev}. In a nutshell, \emph{relative-scale} ambiguity refers to the phenomenon where the estimated trajectory is ambiguous, and is recovered as a one-parameter family of trajectories relative to the ego-camera. Each dynamic body has a different, uncorrelated relative-scale, which renders the problem \emph{unobservable}~\cite{multibody-in-practice} and degeneracy-laden~\cite{kundu2011,two-view-multibody,Namdev}.
%\change{This incites us to explore learning-based methods in the quest for a pragmatic approach to multi-body monocular SLAM.}
This incites us to explore the usage of static feature correspondences in the environment for ego and dynamic vehicle motion estimation in \emph{metric scale}\footnote{We use the term \emph{metric scale} to denote a coordinate frame in which all distances are expressed in units of metres.}.
% Several factors such as lack of robust perceptual models, poor predictive models of dynamic participants, coupled with the scale-factor ambiguity that plagues monocular SLAM in general, make multi-body SLAM infeasible in practice.

% change{We investigate the benefits (and shortcomings) of our learning models for dynamic objects in a scene. We show that learning-methods enable us to cast the multi-body visual SLAM backend into a well-known multi pose-graph optimization framework.}
We propose a multi pose-graph optimization framework for dynamic objects in a scene, and demonstrate its ability to solve for multiple object motions including ego vehicle \emph{unambiguously}, in a \emph{unified global frame} in \emph{metric scale}.
To the best of our knowledge, this is the first monocular multibody SLAM to represent moving obstacle trajectories in a unified global metric frame, on long real-world dynamic trajectories. The quantitative results presented in Sec.~\ref{sec:results} demonstrate the efficacy of the proposed formulation.

\begin{figure}[!t]
    \centering
    \includegraphics[width=0.95\columnwidth ]{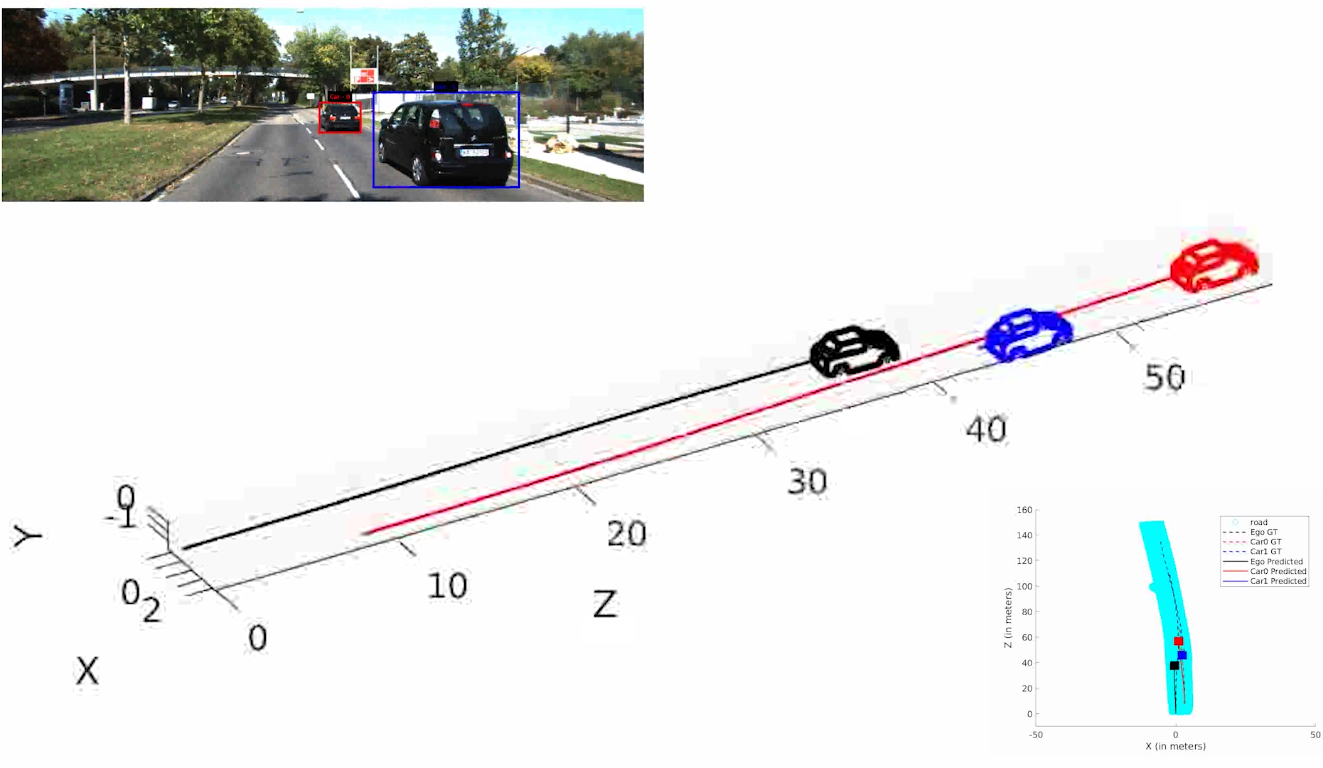}
    \caption{We propose a monocular \emph{multi-object} (multibody) SLAM pipeline which accurately recovers the structure and motion of dynamic participants in the environment in metric scale. Illustrated explanation of the proposed approach and corresponding results can be found \href{https://youtu.be/cchPIaKSSvM}{here}.}
    \label{fig:teaser}
\end{figure}

\begin{figure*}[t]
  \centering
  \vspace{2mm}
  \includegraphics[width=0.98\linewidth]{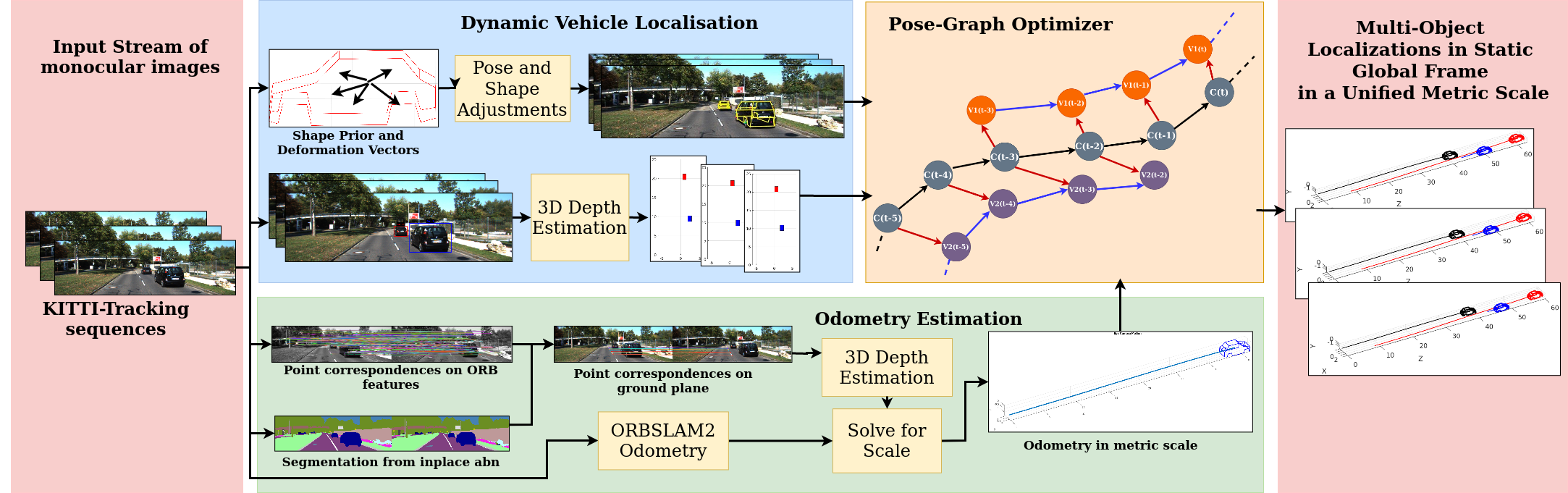}
  \caption{ \textbf{Pipeline}: We obtain dynamic-vehicle localizations via the modules explained in {\color{blue}blue} section. The mathematical representations to the same can be found in \ref{subsec:pspipeline} and \ref{subsec:mobili}. The {\color{green}green} section illustrates our approach to obtain accurate odometry estimations in metric scale, as explained in \ref{subsec:odometry}. The {\color{orange}orange} section illustrates a part of the pose-graph structure where the {\color{gray}gray}, the {\color{orange}orange} and the {\color{purple}purple} nodes represent the nodes for ego-car and two dynamic vehicles in the scene respectively. Moreover, the {\color{black}black}, the {\color{blue}blue} and the {\color{red}red} edges represent the camera-camera, vehicle-vehicle and camera-vehicle edges respectively.}
  \label{fig:pipeline}
\end{figure*}

In the remainder of this paper, we elaborate upon the following key contributions:
\begin{enumerate}
    \item Leveraging single-view metrology for \emph{scale-unambiguous} static feature correspondence estimation
    \item A multi pose-graph formulation that recovers a \emph{metric scale} solution to the multibody SLAM problem.
    % Formulate the relationships between various dynamic vehicles and the ego-vehicle appropriately in the form of a pose-graph.
    % \item Perform localization and mapping of multiple dynamic vehicles in scene and dynamic ego-vehicle using a single optimization problem.
    \item \emph{Practicality}: Evaluation of our approach on challenging sequences from the KITTI driving dataset \cite{kitti}
    % , and present the first \emph{practical} monocular multi-body SLAM system. %We demonstrably fare superior to current multibody SLAM approaches.
\end{enumerate}

\section{Related Work}

The earliest approaches to monocular multibody SLAM~\cite{multibody-kanade,multibody-sam,two-view-multibody, multiple-motion-uncalib,multi-body-seg} were based on \emph{motion segmentation}: segmenting multiple motions from a set of triangulated points. Extending epipolar geometry to multiple objects, \emph{multibody fundamental matrices} were used in \cite{two-view-multibody,two-view-outlier,two-view-robust,multi-body-seg}.

\emph{Trajectory triangulation} methods~\cite{trajectory-triang,multibody_generic}, on the other hand, derive a set of constraints for trajectories of objects, and solve the multi-body SLAM problem under these constraints.
Ozden \etal{} \cite{multibody-in-practice} extend the multi-body Structure-from-Motion framework \cite{multibody-sam} to cope with practical issues, such as a varying number of independently moving objects, track failure, etc.
Another class of approaches applies model selection methods to segment independently moving objects in a scene, and then explicitly solve for relative scale solutions~\cite{two-view-outlier,multibody-sam,multibody-kanade}.
It is worth noting that the above approaches operate \emph{offline}, and extending them for online operation is non-trivial.

Kundu~\etal{}~\cite{kundu2011} proposed a fast, incremental multi-body SLAM system that leverages motion segmentation to assign feature tracks to dynamic bodies, and then independently for relative-scale for the segmented motions. Critical to their success is the underlying assumption of smooth camera motions. Later Namdev~\etal{}~\cite{Namdev} provided analytical solutions for a restricted set of vehicle motions, (linear, planar, and nonholonomic).
% The underlying principle here is that since the problem of computing relative scale is fundamentally ill-posed it can be computed with restrictions on motions of the dynamic participants in the scene. 

More recently, Ranftl~\etal{}~\cite{vibhav2016} presented a dense monocular depth estimation pipeline targeted at dynamic scenes, and resolve \emph{relative scale} ambiguity.
% Using optical flow and motion segmentation, they formulate a convex optimization problem to jointly estimate pixel correspondences across two-views. However, the obtained solution is still scale-ambiguous.
CubeSLAM~\cite{8708251} proposes an object SLAM framework for road scenes. However, it only estimates a \emph{per-frame} relative pose for each object, and does not unify it to construct a trajectory (to avoid relative-scale-ambiguity).
% However, relative scale problems still persist. The authors tackle this ambiguity by defining \emph{support} priors between an object and the environment. Furthermore, the relative-scale resolved reconstruction is not in metric scale. The authors in Cubeslam\cite{8708251} propose an object SLAM framework for on-road scenes. However the problem is not cast into a Dynamic SLAM/ Multibody SLAM context and hence unlike the present work, Cubeslam\cite{8708251} does not show trajectories of dynamic participants and benchmark it with respect to ground truth.

With the advent of deep learning, improvements to object detection~\cite{rota2018place, he2017mask, girshick2015fast, ren2015faster} and motion segmentation have resulted in such methods directly being employed in multi-body SLAM. Reddy~\etal{}~\cite{dinesh2016} and Li~\etal{}~\cite{li2018stereo} present approaches to multi-body SLAM using a stereo camera. In this case, however, the problem is \emph{observable}, while we handle the harder, \emph{unobservable} case of monocular cameras. % [cite paper from Ian Reid's group??]

\section{Overview of the Proposed Pipeline}
\label{sec:pipeline}

With sequence of traffic scene frames as input, our formulation estimates:

\begin{enumerate}
    \item Ego-motion obtained as the camera motion in metres in a static global system for each input frame with an SE(3) formulation.
    \item Trajectory estimates to each object in the traffic scene being captured by the camera in metres in static global frame for each input frame with an SE(3) formulation. 
\end{enumerate}

We obtain the above estimates with a pipeline summarized in the following manner:

\begin{enumerate}
    \item We take a stream of monocular images as input to our pipeline.
    \item We exploit 3D depth estimation to ground plane points as a source of vehicle localizations in ego-camera frame as explained in Sec.~\ref{subsec:mobili}.
    \item Alternatively, as explained in Sec.~\ref{subsec:pspipeline}, we fit a base \emph{shape prior} to each vehicle instance uniquely to obtain refined vehicle localizations in ego-camera frame.
    \item To obtain accurate odometry estimations, we exploit depth-estimates to unique point-correspondences to scale ORB-SLAM2\cite{mur2017orb} (we use ORB-SLAM2 and ORB interchangeably unless otherwise specified) initialization to metric units as explained in Sec.~\ref{subsec:odometry}.
    \item Finally, our optimization formulation (\cf{} Sec.~\ref{sec:optimization}) utilizes the above estimates to resolve \emph{cyclic-consistencies} in the pose-graph.
    \item This provides us with accurate \textbf{multi-body localizations} in a \textbf{static global frame} and consistent \textbf{metric scale}.
\end{enumerate}

\section{Vehicle Localization and Odometry Estimation}
\label{sec:dataPrep}

% \Madhav{A pipeline section after related work giving the overview of each module is critical }

\subsection{Depth Estimation for Points on Ground Plane}
\label{subsec:mobili}

We utilize the known camera intrinsic parameters $K$, ground plane normal $n$, 2D bounding boxes\cite{Redmon_2017_CVPR} and camera height $h$ in metric unit to estimate the depth of any point on the ground plane\footnote{\emph{Flat-earth assumption}: For the scope of this paper, we assume that the autonomous vehicle is operating within a bounded geographic area of the size of a typical city, and that all roads in consideration are \emph{somewhat} planar, i.e., no steep/graded roads on mountains.}. Given the 2D homogeneous coordinates to the point in image space to be $x_{t}$, we estimate the 3D depth to them using the following method as shown in Song \etal{}\cite{song2015joint}. 

\small
\begin{equation}
X_{t} = \frac{- hK^{-1}x_{t}}{n^{T}K^{-1} x_{t}}
\label{eqn:mobiliFormula}
\end{equation}
\normalsize

\subsection{Odometry Estimations}
\label{subsec:odometry}

The initializations to our odometry pipeline (\cf{} Fig.~\ref{fig:pipeline}) come from the ORB trajectory\cite{mur2017orb} in a static-global frame but in an ambiguous ORB scale as opposed to our requirement of \emph{metric scale}. 

We scale the ego-motion from the ORB-SLAM2\cite{mur2017orb} input by minimizing the re-projection error of the ground point correspondences between each pair of consecutive frames. Given frames $t - 1$ and $t$, we have odometry initializations in 3D in ORB scale from ORB-SLAM2 as $T_{t-1}$ and $T_{t}$ respectively. We obtain the relative odometry between the two frames as follows\footnote{We use $\times$ to denote matrix multiplication for the scope of this paper.}:

\small
\begin{equation}
T_{t}^{t-1} = (T_{t-1})^{-1} \times T_{t}
\label{eqn:relativeTransform}
\end{equation}
\normalsize

 We now obtain ORB features to match point correspondences between the two frames $t - 1$ and $t$ and use state-of-the-art semantic segmentation network\cite{rota2018place} to retrieve points $x_{t-1}$ and $x_{t}$ that lie on the ground plane. We obtain the corresponding points $X_{t-1}$ and $X_{t}$ in 3D, given the camera height, via Eqn. \ref{eqn:mobiliFormula} as explained in Sec. \ref{subsec:mobili}. To reduce the noise incorporated by the above method, we only consider points within a threshold in depth of T = 12 metres from the camera. Further, we obtain the required scale-factor $\alpha$ that scales odometry from Eqn. \ref{eqn:relativeTransform} via a minimization problem as shown in Eqn. \ref{eqn:closedFormSoln}, the objective function to which is elaborated as Eqn. \ref{eqn:FunctionDefinitionF(a)}: 

\small
\begin{equation}
F(\alpha) = (X_{t-1} - (R^{t-1}_{t} \times X_{t} + \alpha \mathbf{tr}^{t-1}_{t}))
\label{eqn:FunctionDefinitionF(a)}
\end{equation}
\normalsize

\small
\begin{equation}
% \displaystyle{\min_{\alpha} \left\| X_{t-1} - (R^{t-1}_{t} \times X_{t} + \alpha*T^{t-1}_{t})\right\|_{2}}
% \displaystyle{\min_{\alpha}(X_{t-1} - (R^{t-1}_{t} \times X_{t} + \alpha*T^{t-1}_{t}))^{T} \times (X_{t-1} - (R^{t-1}_{t} \times X_{t} + \alpha*T^{t-1}_{t}))}
\displaystyle{\min_{\alpha}F(\alpha)^{T}\times F(\alpha)}
\label{eqn:closedFormSoln}
\end{equation}
\normalsize
Here, $R_{t}^{t-1}$ and $\mathbf{tr}_{t}^{t-1}$ represent the relative rotation matrix and translation vector respectively. After solving the above minimization problem, we finalize our scale factor $\alpha$ as the mean of solutions obtained from the following:

\small
\begin{equation}
\alpha = \frac{(X_{t-1} -  (R^{t-1}_{t}\times X_{t}))^{T} \times \mathbf{tr}_{t}^{t-1}}{(\mathbf{tr}_{t}^{t-1})^{T} \times \mathbf{tr}_{t}^{t-1}}
\label{eqn:alphaSoln}
\end{equation}
\normalsize

\subsection{Pose and Shape Adjustments Pipeline}
\label{subsec:pspipeline}

We obtain object localizations by using a method inspired from Murthy \etal{}\cite{murthy2017shape}. Our object representation follows from \cite{murthy2017reconstructing, murthy2017shape, ansari2018earth}, based on a \textit{shape prior} consisting of $k$ ordered keypoints which represent specific visually distinguishable features on objects which primarily consist of cars and minivans. For this work, we stick with the 36 keypoints structure from Ansari \etal{}\cite{ansari2018earth}. We obtain the keypoint localizations in 2D image space using a CNN based on stacked hourglass architecture\cite{newell2016stacked} and use the same model trained on over 2.4 million rendered images for Ansari \etal{}\cite{ansari2018earth}.

Borrowing the notations from Murthy \etal{}\cite{murthy2017shape}, we begin with a basis \textit{shape prior} for the object used as a mean shape $\overline{X} \in \mathbb{R}^{3k}$. Let $B$ basis vectors be $V \in \mathbb{R}^{3k \times B}$ and the corresponding deformation coefficients be $\Lambda \in \mathbb{R}^{B}$. Assuming that a particular object instance has a rotation of $R \in SO(3)$ and translation of $\mathbf{tr} \in \mathbb{R}^{3}$ with respect to the camera, its instance $X \in \mathbb{R}^{3k}$ in the scene can be shown mathematically using the following \textit{shape prior} model:

\small
\begin{equation}
X = \hat{R} \times (\overline{X} + V \times \Lambda) + \hat{\mathbf{tr}}
\label{eqn:objectShape}    
\end{equation}
\normalsize
Here, $\hat{R} = diag([R,R,R,...,R]) \in \mathbb{R}^{3k \times 3k}$ and $\hat{\mathbf{tr}} = (\mathbf{tr}^{T},\mathbf{tr}^{T},\mathbf{tr}^{T},...,\mathbf{tr}^{T})^{T} \in \mathbb{R}^{3k}$. Also, $\overline{X} = (\overline{X}_{1}^{T},\overline{X}_{2}^{T},\overline{X}_{3}^{T},...,\overline{X}_{K}^{T}) \in \mathbb{R}^{3k}$ represents the basis \textit{shape prior} and the resultant shape for the object instance is $X = (X_{1}^{T},X_{2}^{T},X_{3}^{T},...,X_{k}^{T}) \in \mathbb{R}^{3k}$ where each $X_{i}$ represents one of the k = 36 keypoints in 3D coordinate system from camera's perspective. Now, Let the ordered collection of keypoint localizations in 2D image space be $\hat{x} = (\hat{x}_{1}^{T},\hat{x}_{2}^{T},\hat{x}_{3}^{T},...,\hat{x}_{k}^{T},) \in \mathbb{R}^{2k}$. Given that $\pi_{k}$ represents the function to project 3D coordinates onto 2D image space using the camera intrinsics $\mu = (f_{x},f_{y},c_{x},c_{y})$, fairly accurate estimates for the pose parameters (R, tr) and the shape parameter ($\Lambda$) for the object instance can be obtained using the following objective function:

\small
\begin{equation}
\min_{R,t,\Lambda} \mathcal{L}_{r} = \left\| \pi_{k}(\hat{R} \times (\overline{X} + V \times \Lambda) + \hat{\mathbf{tr}};f_{x},f_{y},c_{x},c_{y}) - \hat{x} \right\|_{2}^{2}
\label{eqn:2dReprojectionError}
\end{equation}
\normalsize

\small
\begin{equation}
\pi([X,Y,Z]^{T},\mu) = 
\begin{pmatrix}
\frac{f_{x}X}{Z} + c_{x} \\ 
\frac{f_{y}Y}{Z} + c_{y} 
\end{pmatrix} \\
% \pi_{K}((X_{1}^{T},...,X_{K}^{T})^{T},\mu) = (\pi(X_{1}^{T},\mu)^{T},...,\pi(X_{K}^{T},\mu)^{T})^{T}
\label{eqn:pinHoleCamModel}
\end{equation}
\normalsize

Minimizing the objective function (\cf{} Eqn \ref{eqn:2dReprojectionError}) separately for \textit{pose parameters} (R, tr) and \textit{shape parameters} ($\Lambda$) provides us with an optimal fitting of the \textit{shape prior} over the dynamic object. We obtain the object orientation as $R$ after \textit{pose parameter adjustments}. The object's 3D coordinates from the camera $\mathbf{tr}'$ is obtained from the mean of wheel centres.

\section{Multi-Object Pose Graph Optimizer}
\label{sec:optimization}

% \change{odometry, shape pose, chandrakars, pose graph optimization (g2o loop closure)}

\begin{figure}[t]
  \centering
    \vspace{2mm}  
  \includegraphics[width=0.8\linewidth]{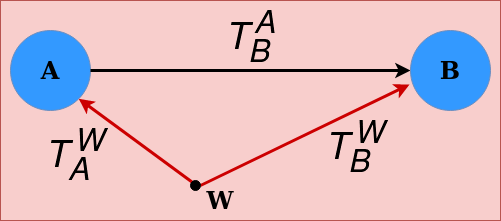}
  \caption{Illustration of a simple pose-graph defined by a constraint defined from nodes A to B by a binary edge.}
  \label{fig:pgSingleEdge}
\end{figure}

\subsection{Pose-Graph Components}
\label{subsec:graphStructure}

Fig. \ref{fig:pgSingleEdge} illustrates a simple pose-graph structure containing two nodes A and B and a binary-edge between them. Using the terminologies from g2o\cite{grisetti2011g2o}, any node A in the pose graph is characterized by a pose $T_{A}^{W} \in SE(3)$ called the \textit{estimate} which defines its pose with respect to the static-global frame of reference W. Meanwhile, a binary-edge from A to B is represented with a relative pose $T_{B}^{A} \in SE(3)$ called the \textit{measurement} which defines the pose of node B from node A's perspective. Mathematically, the constraint introduced by the binary-edge is given as: 

\small
\begin{equation}
\Upsilon_{AB} = (T_{B}^{A})^{-1} \times (T_{A}^{W})^{-1} \times T_{B}^{W}
\label{eqn:g2oCost}
\end{equation}
\normalsize

Assuming relative correctness between each term in Eqn. \ref{eqn:g2oCost}, it results in an identity matrix $I_{4} \in SE(3)$ irrepective of the order of transformation. Thus, Eqn. \ref{eqn:g2oCost} reduces to:

\small
\begin{equation}
T_{A}^{B} \times T_{W}^{A} \times T_{B}^{W} = I_{4}
\label{eqn:simplifiedG2oCost}
\end{equation}
\normalsize

Clearly, the order in which the transformations are applied do not change the consistency of the respective cycle in the pose-graph. Thus, Eqn. \ref{eqn:simplifiedG2oCost} can also be written as:

\small
\begin{equation}
T_{B}^{W} \times T_{A}^{B} \times T_{W}^{A} = I_{4}
\label{eqn:secondSimplified}
\end{equation}
\normalsize

\subsection{Pose-Graph Formulation}
\label{subsec:costs}

% As explained in \ref{subsec:graphStructure}, given that $W$ represents the static world coordinate frame of reference, nodes A and B are parameterized by poses $T_{A}^{W} \in SE(3)$ and $T_{B}^{W} \in SE(3)$ respectively, and a binary edge from node A to node B is parameterized using a pose $T_{B}^{A} \in SE(3)$. Each binary edge between any pair of nodes represents a unique constraint between the respective nodes. This constraint can be mathematically represented as follows:

\begin{figure}[t]
  \centering
  \vspace{1.7mm}
  \includegraphics[width=0.60\linewidth]{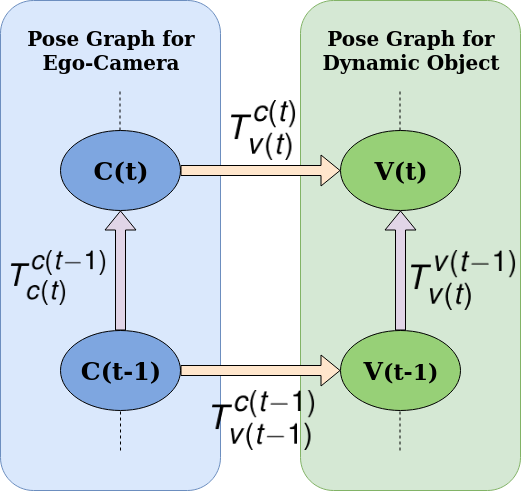}
  \caption{Illustration of our multi-body pose-graph structure defined between a pair of consecutive frames. Nodes in {\color{blue} blue} correspond to the primary pose-graph for the ego-motion while those in {\color{green} green} correspond to the secondary pose-graph for the dynamic objects in the scene.}
  \label{fig: main_architecture}
\end{figure}

% \small
% \begin{equation}
% \Upsilon_{AB} = (T_{B}^{A})^{-1} \times (T_{A}^{W})^{-1} \times T_{B}^{W}
% \label{eqn:g2oCost}
% \end{equation}
% \normalsize

\begin{figure*}[h]
    \centering
    \vspace{2mm}
    \begin{adjustbox}{max width=\linewidth}
    \begin{tabular}{ccc}
    
        \includegraphics[width = 0.25\linewidth]{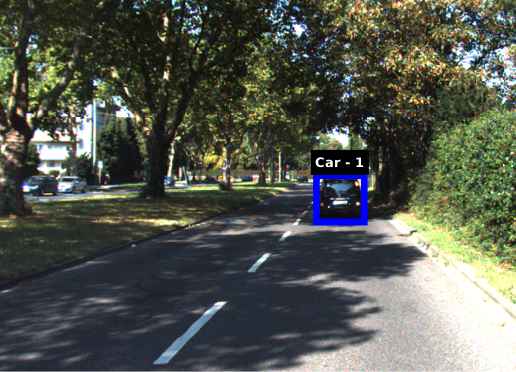}
        & \includegraphics[width = 0.5\linewidth, height = 40mm]{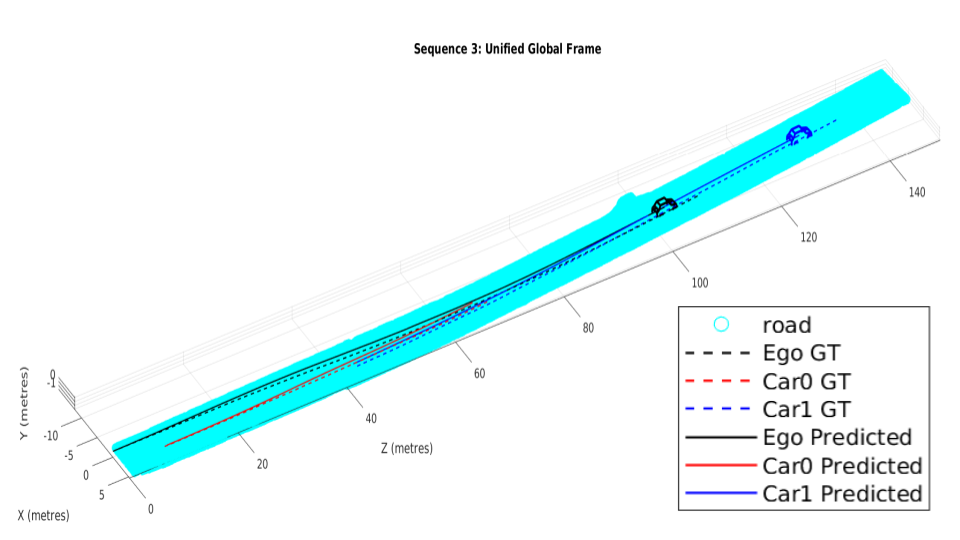}
        & \includegraphics[width = 0.25\linewidth, height = 40mm]{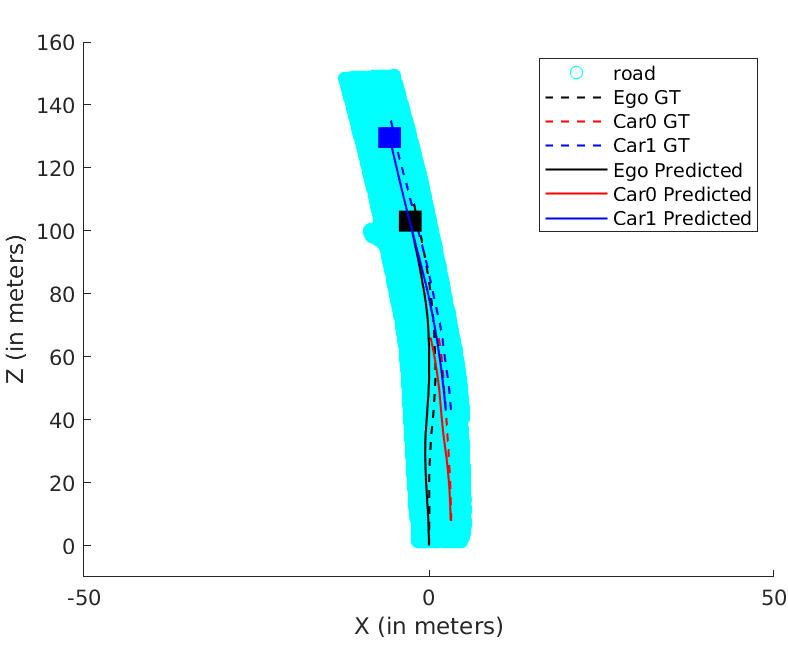} \\
        
        \includegraphics[width = 0.25\linewidth]{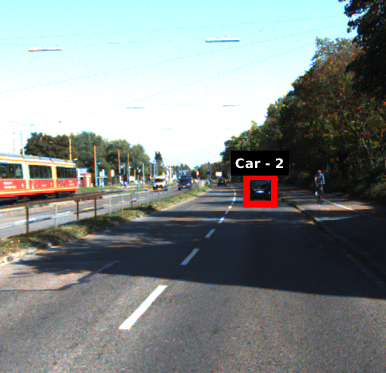} 
         & \includegraphics[width = 0.5\linewidth, height = 40mm]{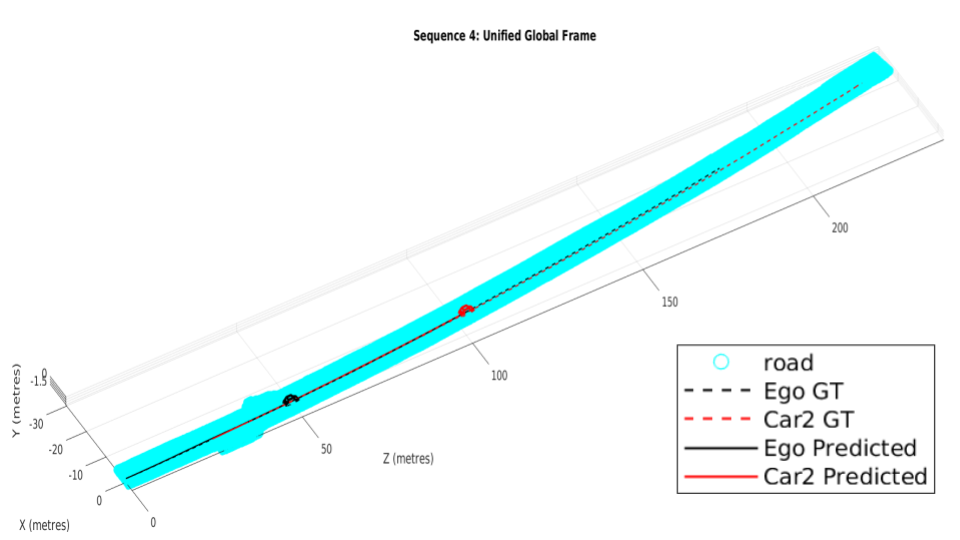}
         & \includegraphics[width = 0.25\linewidth, height = 40mm]{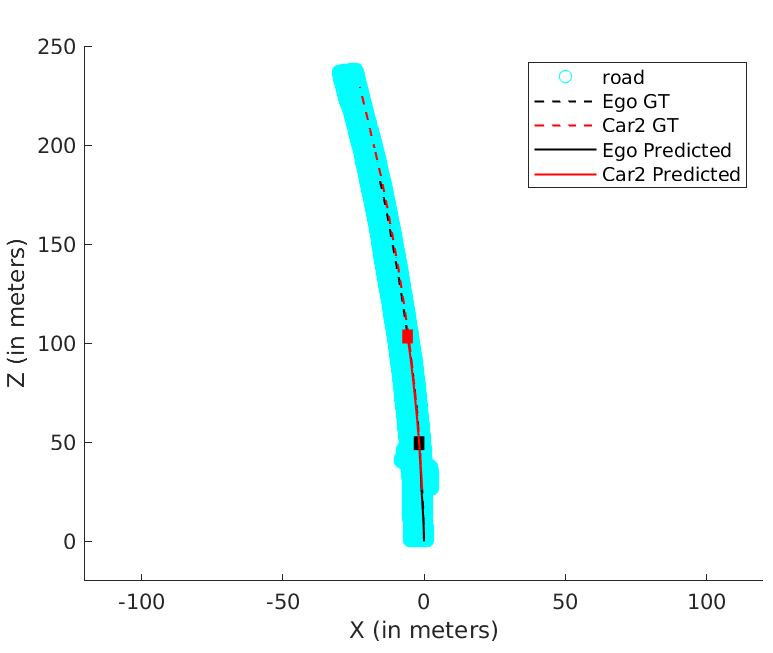} \\
         
          \includegraphics[width = 0.25\linewidth]{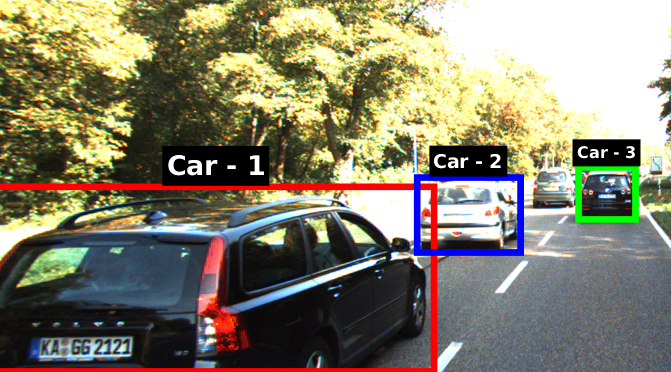}
         & \includegraphics[width = 0.5\linewidth, height = 40mm]{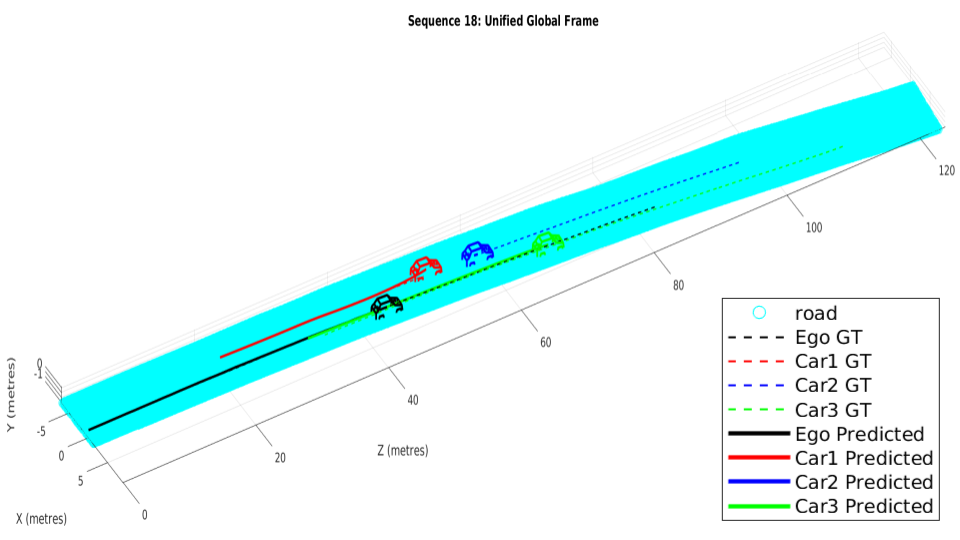}
         & \includegraphics[width = 0.25\linewidth, height = 40mm]{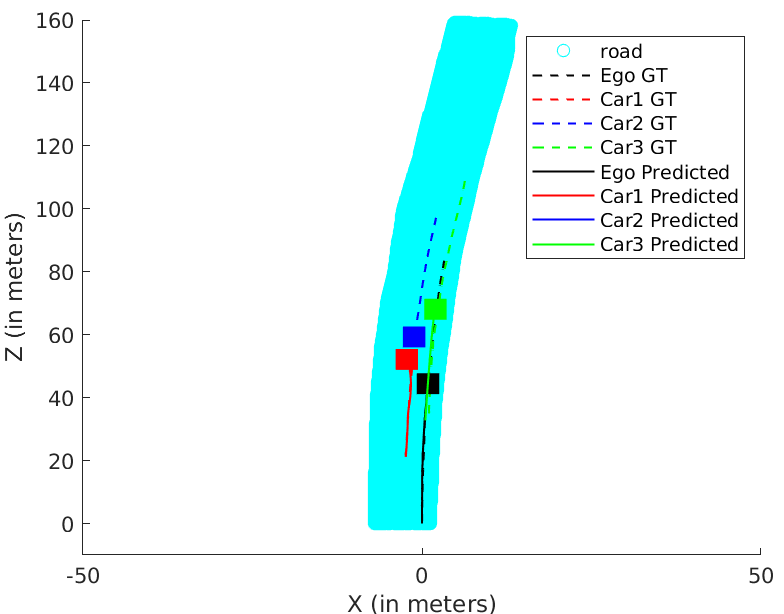} \\
        
    \end{tabular}
    \end{adjustbox}
    \caption{Qualitative results on various sequences. \emph{Col 1} shows the input images with bounding boxes to specify the vehicles mapped in \emph{Col 2} and \emph{Col 3}. While \emph{Row 1} and \emph{Row 3} illustrate our performance on multi-vehicle road plane scenarios, \emph{Row 2} shows results for a far away vehicle over a long sequence. Ego-vehicle is shown in black whereas the {\color{red}red}, {\color{blue}blue} and {\color{green}green} plots represent the unique vehicle instances in the scene with the corresponding dotted plots showing the ground truths. Note that the entire ground truth trajectory is shown at once in the figures whereas the predicted trajectories are up to the instance frame shown in \emph{Col 1}. More detailed results can be found \href{https://youtu.be/cchPIaKSSvM}{here}.}
    \label{fig:qualitative_results}
\end{figure*}

% Assuming relative correctness between each term in Eqn. \ref{eqn:g2oCost}, it results in an identity matrix $I_{4} \in SE(3)$ irrepective of the order of transformation. Thus, Eqn. \ref{eqn:g2oCost} reduces to:

% \small
% \begin{equation}
% T_{A}^{B} \times T_{W}^{A} \times T_{B}^{W} = I_{4}
% \label{eqn:simplifiedG2oCost}
% \end{equation}
% \normalsize

% We can see that the order in which the transformations are applied do not change the consistency of the respective cycle in the pose-graph. Thus, Eqn. \ref{eqn:simplifiedG2oCost} can also be written as follows:

% \small
% \begin{equation}
% T_{B}^{W} \times T_{A}^{B} \times T_{W}^{A} = I_{4}
% \label{eqn:secondSimplified}
% \end{equation}
% \normalsize

Fig. \ref{fig: main_architecture} illustrates the pose graph structure between every consecutive set of frames $t - 1$ and $t$ containing four nodes and four edges between them. We obtain the \textit{estimates} for camera nodes $c(t-1)$ and $c(t)$ (i.e., $T_{c(t-1)}^{W}$ and $T_{c(t)}^{W}$) and \textit{measurement} for the camera-camera edge (i.e. $T_{c(t)}^{c(t-1)}$) from our odometry estimation (\cf{} Sec. \ref{subsec:odometry}). We use this odometry to register dynamic object localizations from pose-shape adjustment pipeline as explained in Sec. \ref{subsec:pspipeline} to provide for the \textit{estimates} $T_{v(t-1)}^{W}$ and $T_{v(t)}^{W}$ to vehicle nodes $v(t-1)$ and $v(t)$. We obtain \textit{measurement} for the camera-vehicle edge (i.e., $T_{c(t-1)}^{v(t-1)}$, $T_{c(t)}^{v(t)}$) from shape and pose adjustment (\cf{} Sec. \ref{subsec:pspipeline}). Moreover, we use depth estimation from ground plane using Song \etal{}\cite{song2015joint} as explained in Sec. \ref{subsec:mobili} as a source of vehicle localization that is unique from the localizations obtained from Sec. \ref{subsec:pspipeline}. This registered with our odometry estimations provides for our vehicle-vehicle edge \textit{measurement} i.e., $T_{v(t)}^{v(t-1)}$. Now, from Eqn. \ref{eqn:simplifiedG2oCost}, the cost function for the above binary-edges, $\Upsilon_{cc}$, $\Upsilon_{cv(t-1)}$, $\Upsilon_{cv(t)}$ and $\Upsilon_{vv}$, can be defined mathematically as:

\small
\begin{equation}
\begin{split}
\Upsilon_{cc} & = T_{c(t-1)}^{c(t)} \times T_{W}^{c(t-1)} \times T_{c(t)}^{W} \\ 
\Upsilon_{cv(t-1)} & = T_{c(t-1)}^{v(t-1)} \times T_{W}^{c(t-1)} \times T_{v(t-1)}^{W} \\
\Upsilon_{cv(t)} & = T_{c(t)}^{v(t)} \times T_{W}^{c(t)} \times T_{v(t)}^{W} \\
\Upsilon_{vv} & = T_{v(t-1)}^{v(t)} \times T_{W}^{v(t-1)} \times T_{v(t)}^{W} 
\label{eqn:indCost}
\end{split}
\end{equation}
\normalsize

Cumulatively, the above cost functions for a single loop illustrated in Fig. \ref{fig: main_architecture} can be represented as:

\small
\begin{equation}
\Upsilon = \Upsilon_{cc} \times \Upsilon_{cv(t)} \times (\Upsilon_{vv})^{-1} \times (\Upsilon_{cv(t-1)})^{-1}
\label{eqn:cumulCost}
\end{equation}
\normalsize

On substituting Eqn. \ref{eqn:indCost} in Eqn. \ref{eqn:cumulCost}, and on further simplification, we obtain the resultant function for cumulative cost which clearly defines the cyclic consistency within the loop defined by the four binary edges: 

\small
\begin{equation}
\Upsilon = T_{c(t)}^{c(t-1)} \times T_{v(t)}^{c(t)} \times T_{v(t-1)}^{v(t)} \times T_{c(t-1)}^{v(t-1)} = I_{4}
\label{eqn:cumulCostSimplified}
\end{equation}
\normalsize

\subsection{Confidence Parameterization}
\label{subsec:infoMatScaling}

In addition to the relative pose between the participating graph nodes, the parameterization for each edge also includes a positive semi-definite inverse covariance matrix or the information matrix $\Omega_{E} \in \mathbb{R}^{N \times N}$ where E represents an edge in the pose graph and N represents the dimension of the Lie group in which the poses are defined. In this work, all poses and transformations are defined in SE(3), hence we can take N = 6 for the information matrix $\Omega_E$ corresponding to each edge E in the whole pose graph. We utilize this as a confidence parameterization for the sources of input-data into the pose-graph. To make the most out of this, we scale the information matrix for an edge $E$ by a scale factor $\lambda \in \mathbb{R}$ to get the effective information matrix $\overline{\Omega}_{E}$ that is finally sent as a parameter:

\small
\begin{equation}
\overline{\Omega}_{E} = \lambda  \Omega_{E}
\label{eqn:infoScaling}
\end{equation}
\normalsize

We categorize all the edges in our pose-graph formulation into three types namely camera-camera, camera-vehicle, and vehicle-vehicle edges. Each type of edges corresponds to a unique source of data to provide for the corresponding constraint. This formulation coupled with the corresponding \emph{confidence parameter} $\lambda$, enables us to scale the effects of the respective categories of edges appropriately. Given that odometry estimates are fairly reliable, we assign a relatively high constant scaling to its information matrix for our experiments on all sequences.

Given that we obtain dynamic vehicle localizations in camera frame from two different sources as explained in Sec. \ref{subsec:graphStructure}, we make intelligent use of the confidence parameter $\lambda$, to scale the information matrix corresponding to the camera-vehicle and the vehicle-vehicle edges in our pose graph. It has been observed over a large number of vehicles that localizations obtained from Sec. \ref{subsec:pspipeline} performs better than the those obtained from Sec. \ref{subsec:mobili} for vehicles closer to the camera (up to about 45 metres). This can be attributed to the keypoint localizations being inaccurate for far away objects. However, estimates from Sec. \ref{subsec:mobili} are more accurate at depths far away from the camera (over 45 metres). Factors like visible features on vehicles do not affect these estimates.

% \Madhav{So what does this scaling achieve or handle the complementary nature of the two depth estimation modules? This needs further explanation}

\section{Experiments and Results}
\label{sec:results}

\subsection{Dataset} 

We test our procedure over a wide range of KITTI-Tracking training sequences\cite{kitti}, spanning over rural and urban scenarios with various number of dynamic objects in the scene. We perform localizations on objects primarily consisting of cars and mini-vans. Our localization pipeline provides accurate results over objects irrespective of the direction of motion and maneuvers undertaken by both the ego-car and the vehicles in the scene. The labels provided in the dataset are used as ground truth for getting the depth to the vehicle's center from the camera. The corresponding ground truth for odometry comes from GPS/IMU data, which is compiled using the OXTS data provided for all the KITTI-Tracking training sequences.

\subsection{Qualitative Results}

\subsubsection{Pose and Shape Adjustments}

\begin{figure}[H]
    %\vspace{2mm}
    \centering
    \begin{adjustbox}{max width=\linewidth}
    \begin{tabular}{cc}
        \raisebox{0.3\height}{\includegraphics[width=0.6\linewidth]{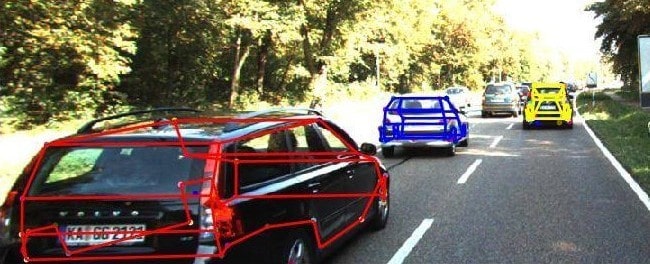}} & 
        \includegraphics[width=0.4\linewidth]{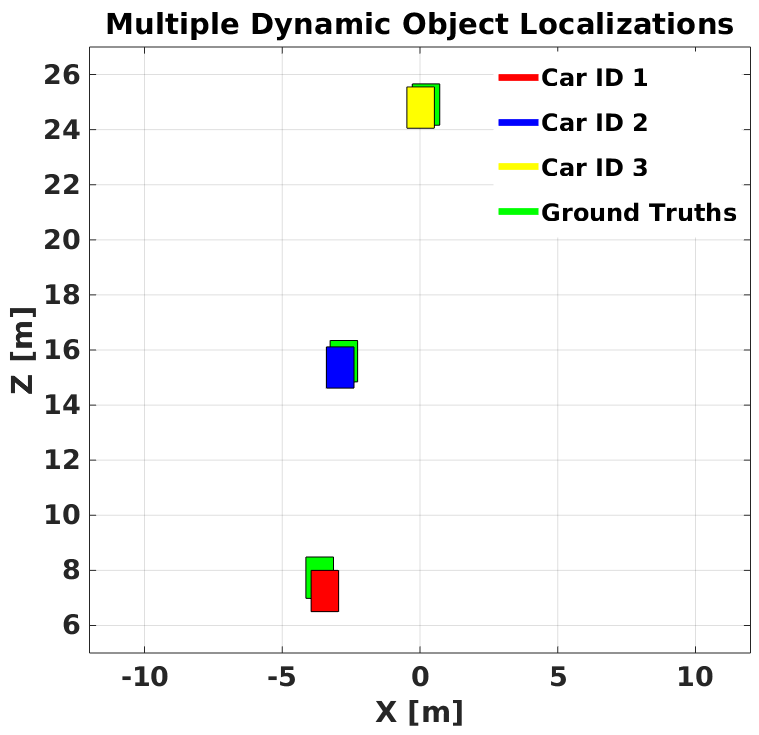}
    \end{tabular}
    \end{adjustbox}
    \caption{Localizations in ego-camera frame after pose and shape estimations for a dynamic multi-vehicle scenario.}
    \label{fig: pose_shape_figure}
\end{figure}

We obtain accurate localizations in ego-camera frame by fitting base \emph{shape priors} to each non-occluded and non-truncated vehicle in the scene with respect to the ego-camera. While the pipeline is dependent on the keypoint localizations on these vehicles, factors like large depth from camera are bound to affect the accuracies with respect to ground truth. However, this approach ensures fairly accurate vehicle localizations for the pose-graph optimizer to apply its edge constraints. Fig. \ref{fig: pose_shape_figure} illustrates wireframe fitting and subsequent mapping in ego-frame for a traffic scenario consisting of multiple vehicles.

\subsubsection{Odometry Estimation}

\begin{figure}[H]
    \centering
    \begin{adjustbox}{max width=\linewidth}
    \begin{tabular}{cc}
        \includegraphics[width=0.5\linewidth]{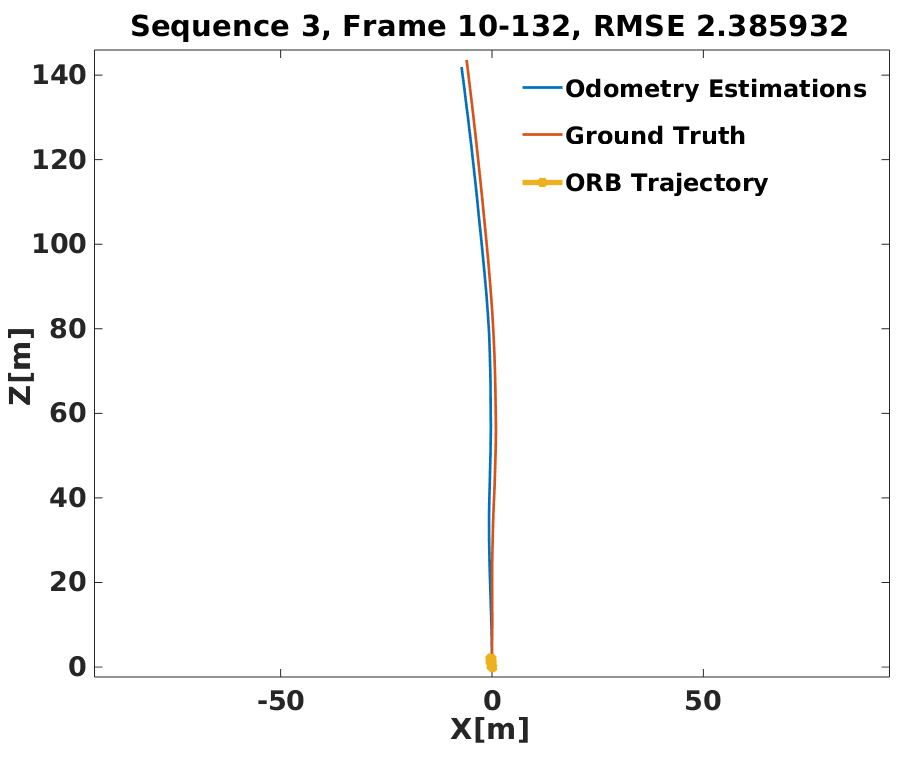}
        & \includegraphics[width=0.5\linewidth]{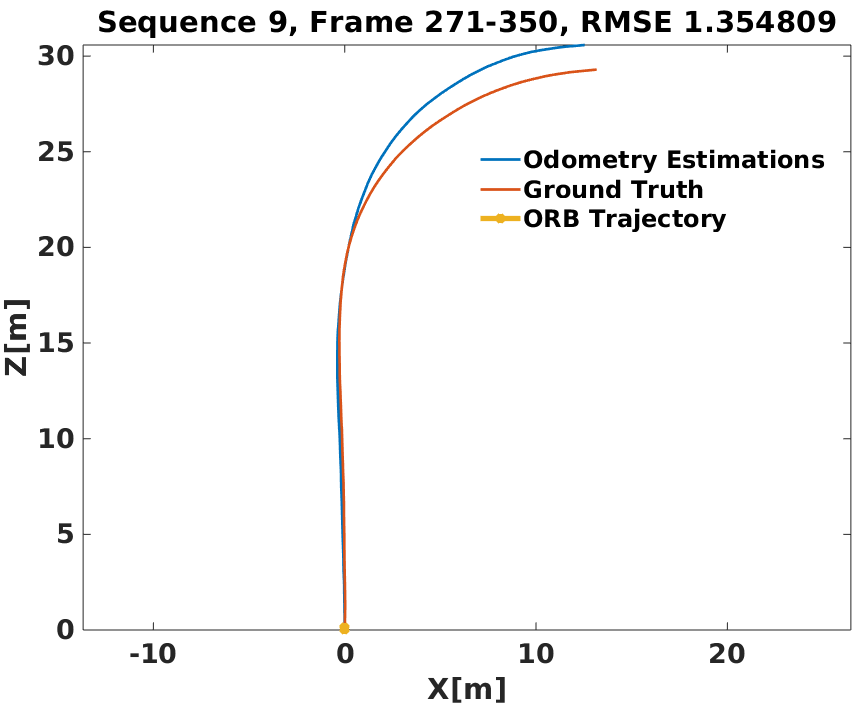}
    \end{tabular}
    \end{adjustbox}
    \caption{Odometry estimations in metric scale in {\color{blue}blue}. GPS/IMU trajectory is indicated in {\color{red}red} and ORB trajectory in its scale is indicated in {\color{yellow}yellow}. The figure illustrates that our method for estimating odometry is proficient on sharp turns and long sequences.}
    \label{fig: odometry_figure}
\end{figure}

For accurate visual odometry, we exploit distinguishable static ORB\cite{orb, mur2017orb} features on the road plane from entities like curbs, lane markers and any irregularities on the road to obtain quality point correspondences. While the approach is dependent on factors like reasonable visibility, we obtain robust performance over a diverse range of sequences many of which are over a 100 frames long. Fig. \ref{fig: odometry_figure} illustrates how our method achieves a fairly accurate scaling of odometry to provide an initialization that competes well with the corresponding ground truth. 

\subsubsection{Pose-Graph Optimization}

We resolve for each cyclic-loop created by the ego-camera and each vehicle (\cf{} Eqn. \ref{eqn:cumulCostSimplified}) in the scene in our optimization formulation. The optimization problem runs for a maximum of 100 iterations. Our optimization formulation performs consistently well on a wide range of sequences irrespective of the sequence length, number of objects in the scene and varying object instance lengths. A unique pose-graph structure for all vehicles including ego-motion at each time instance ensures effective error re-distribution across all trajectories based on efficient confidence parameterization (\cf{} Eqn. \ref{eqn:infoScaling}). Fig. \ref{fig:qualitative_results} illustrates ego-motion as well as the motion of various vehicles over many sequences from the KITTI Tracking dataset\cite{kitti}.

\subsection{Quantitative Results}

\subsubsection{Odometry Estimations}
As an attempt to improve odometry estimations, we place a threshold T on the depth from camera upto which we consider point correspondences. This has been set based on our observation that the accuracy of the 3D depth to the point correspondences lowers with depth from the camera. Table \ref{table:ego_vehicle_error} summarizes our experiments with various threshold values before we finalize our threshold at T = 12 metres.

 %---------- ODOMETRY TABLE ---------------%
\begin{table}[H]
\begin{center}
\begin{adjustbox}{max width=\textwidth}
\begin{tabular}{|c|c||c|c|c|c|}

\hline
\multirow{2}{*}{Seq no.} & \multirow{2}{*}{Seq length} & \multicolumn{4}{|c|}{Threshold (metres)} \\ \cline{3-6}

& & $12$ & $15$ & $18$ & $20$ \\ \hline 

$1$ & $41$ & $\mathbf{4.39}$ & $5.63$ & $5.61$ & $5.18$ \\ \hline 

% $2$ & $111$ & $8.58$ & $6.71$ & $7.02$ & {$\mathbf{5.93}$} \\ \hline 

$3$ & $123$ & {$\mathbf{1.65}$} & $2.45$ & $1.91$ & $2.57$ \\ \hline

$4$ & $149$ & {$\mathbf{7.64}$} & $8.84$ & $9.59$ & $10.96$ \\ \hline 

$6$ & $51$ & $5.90$ & {$\mathbf{2.37}$} & $2.38$ & $2.82$ \\ \hline

$9$ & $80$ & $5.52$ & {$\mathbf{1.35}$} & $1.44$ & $1.44$ \\ \hline 

$18$ & $141$ & {$\mathbf{1.98}$} & $3.31$ & $2.98$ & $3.36$ \\ \hline

\multicolumn{2}{|c||}{Average ATE} & $4.51$ & {$3.99$} & $\mathbf{3.98}$ & $4.39$ \\ \hline 

\end{tabular}
\end{adjustbox}
\end{center}
\caption{Analysis between various threshold settings for odometry estimations by computing Absolute Translation Error (ATE) in metres. (\cf{} Sec. \ref{subsec:odometry})}
\label{table:ego_vehicle_error}
\end{table}
%--------------------------------------------------%

%----------- COMPARISON TABLE -----------%
\begin{table*}[t]
\vspace{1.7mm}
\begin{center}
\begin{adjustbox}{max width=\linewidth}
\begin{tabular}{|c|c|c|c|c|c|c|c|c|c|c|}

\hline
& \multicolumn{10}{c|}{Absolute Translation Error (Root Mean Square) in Global Frame (metres)} \\ \hline

Seq No. & \multicolumn{3}{c|}{3} & \multicolumn{2}{c|}{4} & \multicolumn{4}{c|}{18} & \multirow{3}{*}{Avg Error} \\ \cline{1-10}

Car ID & $0$ & $1$ & Ego-car & $2$ & Ego-car & $1$ & $2$ & $3$ & Ego-car &\\ \cline{1-10}

Frame length & $41$ & $92$ & $123$ & $149$ & $149$ & $62$ & $83$ & $141$ & $141$ & \\ \hline

Namdev \etal{}\cite{Namdev} & $13.81$ & $11.58$ & $11.49$ & $11.18$ & $11.12$ & $3.77$ & $5.93$ & $3.72$ & $3.69$ & $8.47$ \\ \hline

Ours & $\mathbf{1.61}$	& $\mathbf{4.99}$ &	$\mathbf{1.96}$ & $\mathbf{2.14}$ &	$\mathbf{6.49}$	& $\mathbf{1.29}$ & $\mathbf{3.45}$	& $\mathbf{2.40}$ & $\mathbf{2.27}$ & $\mathbf{2.96}$ \\ \hline

\end{tabular}
\end{adjustbox}
\end{center}
\caption{Comparative performance based on ATE of our pipeline.}
\label{table:comptable}
\end{table*}
% %--------------------------------------------------

 %----------- VEHICLE LOCALIZATION TABLE -----------%
\begin{table*}[t]
\vspace{1.7mm}
\begin{center}
\begin{adjustbox}{max width=\linewidth}
\begin{tabular}{|c|c|c|c|c|c|c|c|c|c|c|}

\hline
& \multicolumn{10}{c|}{Absolute Translation Error (ATE) (Root Mean Square) in Global Frame (metres)} \\ \hline

Seq number & \multicolumn{3}{c|}{3} & \multicolumn{2}{c|}{4} & \multicolumn{4}{c|}{18} & \multirow{3}{*}{Avg Error} \\ \cline{1-10}

Car ID & $0$ & $1$ & Ego-car & $2$ & Ego-car & $1$ & $2$ & $3$ & Ego-car &\\ \hline

Frame length & $41$ & $92$ & $123$ & $149$ & $149$ & $62$ & $83$ & $141$ & $141$ & \\ \hline

Initialization & $1.62$ & $4.99$ & $1.96$ & $13.65$ & $6.43$ & $1.33$ & $3.47$ & $3.53$ & $2.24$ & $4.36$ \\ \hline

Only C-C and C-V edges  & $1.62$ & $5.01$ & $1.98$ & $13.65$ & $6.43$ & $1.32$ & $3.48$ & $3.24$ & $2.24$ & $4.33$ \\ \hline

Only C-C and V-V edges  & $2.88$ & $5.22$ & $1.96$ & $2.14$ & $6.43$ & $1.29$ & $4.00$ & $2.80$ & $\mathbf{2.24}$ & $3.22$ \\ \hline

Only C-V and V-V edges  & $1.61$ & $5.68$ & $3.54$ & $2.24$ & {$\mathbf{6.41}$} & $1.65$ & $\mathbf{3.03}$ & $\mathbf{2.24}$ & $2.76$ & $3.23$ \\ \hline

With C-C, C-V and V-V edges  & {$\mathbf{1.61}$} & {$\mathbf{4.99}$} & {$\mathbf{1.96}$} & {$\mathbf{2.14}$} & $6.49$ & $\mathbf{1.29}$ & $3.45$ & $2.40$ & $2.27$ & $\mathbf{3.01}$ \\ \hline

Percentage Errors  & $6.60\%$ & $2.07\%$ & $-$ & $1.23\%$ & $-$ & $3.36\%$ & $3.39\%$ & $2.02\%$ & $-$ & $3.11\%$ \\ \hline

\end{tabular}
\end{adjustbox}
\end{center}
\caption{ATE for all vehicles in a static-global frame recognized by the pose-graph formulation across various sequences. The percentage error (with C-C, C-V and V-V edges) with respect to ground-truth depth explains the drift experienced by the vehicles in the scene with respect to both its total distance traveled and its initial depth from the static global frame. The same is not shown for Ego-Car as the denominator for this metric becomes very small since the ego-motion begins from the global origin.}
\label{table:vehicle_localisation_error}
\end{table*}
%--------------------------------------------------

While T = 12 meters delivers best results for most sequences mentioned in Table \ref{table:ego_vehicle_error}, we see that T = 15 metres performs better for sequences 6 and 9, both of which involve the ego-vehicle taking a sharp turn at an intersection.
 This is because we rely on ground plane features including and largely contributed to by the lane markers on the road plane. Given that the segment of road plane in the scene at an intersection is devoid of any road/lane markers, we do not get enough feature correspondences from closer segments of the road. Meanwhile, increasing the threshold enables us to pick up points from the road plane continuing beyond the intersection which contains better scope for quality feature correspondences in the form of lane markers. Consequently, a relatively larger threshold performs better.

 \subsubsection{Pose-Graph Optimization}
 Our pose-graph formulation consists of three categories of edges namely camera-camera(C-C) edges, camera-vehicle(C-V) edges and Vehicle-Vehicle(V-V) edges. Each of these sets of edges are accompanied by a unique \emph{confidence parameter} $\lambda$. To understand the contribution category of edges to our pose-graph optimization, we analyse the results on removing these constraints. Table \ref{table:vehicle_localisation_error} summarizes our observations. It can be noted that few vehicles in sequence 3 and the ego-vehicle in sequence 4 perform better when C-C constraints are relaxed. This is because, the optimizer generally utilizes reliable edges in each loop of the pose-graph to improve the relatively less reliable edges, provided their information matrices are scaled appropriately. Given that the C-C edges are less reliable in these sequences, relaxing its constraints enables other edges to improve upon the overall error. A similar explanation can be given for the errors for ego-motion in sequence 18. Since C-C edges of the ego-motion in sequence 18 are more accurate than the corresponding C-V edges of other vehicles, we obtain a better result for the same when the C-V edge constraint is relaxed. Both C-C and V-V edges are generated using the odometry estimations and are influenced by its accuracy too.
 
 Table \ref{table:comptable} compares our performance with Namdev \etal{}\cite{Namdev}. Since ATE is not reported in their literature, we calculate the ATE after running the available implementation. As is evident from Table \ref{table:comptable}, we showcase superior performance in all sequences when compared with Namdev \etal{}\cite{Namdev}.

\subsection{Summary of Results}

While Fig. \ref{fig:qualitative_results} illustrates how our trajectories perform with respect to the ground truth, Table \ref{table:vehicle_localisation_error} reaffirms how our pose-graph formulation successfully redistributes errors about constraints with high \emph{confidence parameters}. Table \ref{table:comptable} reports our pipeline's performance with respect to Namdev \etal{}\cite{Namdev}. Tables \ref{table:vehicle_localisation_error} and \ref{table:comptable} vindicate the efficacy of the proposed pipeline as the absolute translation error(ATE) are typically around 3m for sequences more than 100m in length. The last row of Table \ref{table:vehicle_localisation_error} denotes the percentage error, which is significantly low for fairly long sequences at an average of 3.11\%, considering that the original problem is intractable and hard to solve.

%%%%%%%%%%%%%%%%%%%%%%%%%%%%%%%%%%%%%%%%%%%%%%%%%%%%%%%%%%%%%%%%%%%%%%%%%%%%%%%%%%%%%%%%%%%%%%%%%%%%%%%%%%%

\section{Conclusion}
\label{sec:conclusions}

Monocular Multi-body SLAM is ill-posed as it is impossible to triangulate a moving vehicle from a moving monocular camera. This observability problem manifests in the form of relative scale when posed into the Multibody framework. With the arrival of single view reconstruction methods based on Deep Learning, some of these difficulties are alleviated, but one is still entailed to represent the camera motion and the vehicles in the same scale. This paper solves for this scale by making use of the ground plane features thereby initializing the ego vehicle and other dynamic participants with respect to a unified frame in metric scale. Further, a pose graph optimization over vehicle poses between successive frames mediated by the camera motion formalizes the Multibody SLAM framework. 

We showcase trajectories of dynamic participants and the ego vehicle over sequences of more than a hundred frames in length with high fidelity ATE (Absolute Translation Error). To the best of our knowledge, this is the first such method to represent vehicle trajectories in the global frame over long sequences. The pipeline is able to accurately map trajectories of dynamic participants far away from the ego camera and its scalability to map multi-vehicle trajectories is another salient aspect of this work. 

% In this work, we presented a \emph{unique} approach for a \emph{practical} monocular multi-object SLAM which gives dynamic vehicle and ego camera localizations in a \emph{unified} framework in \emph{metric} scale. By doing this, we successfully utilized relationships between each localized entity in a dynamic traffic scenario to improve accuracies at a collective level. Future work could be to improve and build over our proposed pipeline in an attempt to tackle cases of occlusions and truncations successfully in an intelligent manner. 

% \section{Acknowledgements}
% \label{sec:acknowledgements}

% The authors would like to thank J. Krishna Murthy for his contributions into this work in the form of valuable discussions and review. 

%%%%%%%%%%%%%%%%%%%%%%%%%%%%%%%%%%%%%%%%%%%%%%%%%%%%%%%%%%%%%%%%%%%%%%%%%%%%%%%%

\small{
\bibliography{references}
\bibliographystyle{IEEEtran}
}

\end{document}